\def\year{2023}\relax
\pdfoutput=1

\documentclass[letterpaper]{article}
\usepackage{aaai23}

\usepackage{times}  
\usepackage{helvet}  
\usepackage{courier}  
\usepackage[hyphens]{url}  
\usepackage{graphicx} 
\usepackage{natbib}  
\usepackage{caption} 
\DeclareCaptionStyle{ruled}{labelfont=normalfont,labelsep=colon,strut=off} 
\usepackage{algorithm}
\usepackage{algorithmic}

\usepackage{newfloat}
\usepackage{listings}
\floatstyle{ruled}
\newfloat{listing}{tb}{lst}{}
\floatname{listing}{Listing}
\usepackage{booktabs}
\usepackage{tabularx}
\usepackage{subfig}
\usepackage{graphicx}
\usepackage{nicematrix}
\usepackage{multirow}
\usepackage{arydshln}

\title{Removing Non-Stationary Knowledge From Pre-Trained Language Models for Entity-Level Sentiment Classification in Finance}
\author {
    Guijin Son,\textsuperscript{\rm 1}
    Hanwool Lee, \textsuperscript{\rm 2}
    Nahyeon Kang, \textsuperscript{\rm 3}
    Moonjeong Hahm \textsuperscript{\rm 4}
}
\affiliations {
    \textsuperscript{\rm 1} Underwood International College, Yonsei University, Republic of Korea \\
    \textsuperscript{\rm 2} NCSOFT, Republic of Korea\\
    \textsuperscript{\rm 3} KB Kookmin Bank, Republic of Korea\\
    \textsuperscript{\rm 4} College of Business \& Economics, Chung-Ang University, Republic of Korea\\
    spthsrbwls123@yonsei.ac.kr, 
    albertmade@ncsoft.com, 
    kangnahyeonkang@gmail.com,
    daily6298@cau.ac.kr
}

\begin{document}

\maketitle

\begin{abstract}
Extraction of sentiment signals for stock movement prediction from news text, stock message boards, and business reports have been a rising field of interest in finance. Building upon past literature, the most recent work attempt to better capture sentiment from sentences with complex syntactic structures by introducing aspect-level sentiment classification (ASC). Despite the growing interest, however, fine-grained sentiment analysis has not been fully explored in non-English literature due to the shortage of annotated finance-specific data. Accordingly, it is necessary for non-English languages to leverage datasets and pre-trained language models (PLM) of different domains, languages, and tasks to improve their performance. To facilitate finance-specific ASC research in the Korean language, we build \( KorFinASC\), a Korean aspect-level sentiment classification dataset for finance consisting of 12,613 human-annotated samples, and explore methods of intermediate transfer learning. Our experiments indicate that past research has been ignorant towards the potentially wrong knowledge of financial entities encoded during the training phase, which has overestimated the predictive power of PLMs. In our work, we use the term ``non-stationary knowledge'' to refer to information that was previously correct but is likely to change, and present ``TGT-Masking'', a novel masking pattern to restrict PLMs from speculating knowledge of the kind. Finally, through a series of transfer learning with TGT-Masking applied we improve 22.63\%  of classification accuracy compared to standalone models on \( KorFinASC\). 
\end{abstract}

\noindent Over the last decade, sentiment analysis has been widely adopted in the field of finance to extract sentiment signals from news text, stock message boards, and business reports \cite{ding-etal-2014-using, 10.1093/rapstu/raz007}. While it has been expected to automate the extraction of nuances from sentiment-bearing expressions fully, traditional coarse-grained sentiment analysis techniques fail to disambiguate sentences including conflicting sentiments. For example, given the following news headline: “Energy stocks rallied while S\&P 500 falls into a bear market”, FinBERT~\cite{https://doi.org/10.48550/arxiv.1908.10063} fails to differentiate between the two contradictory sentiments present  . To resolve such problems, recent studies introduce aspect-level sentiment classification (ASC), a task that identifies the polarity of each sentence with specific entities in interest \cite{CONSOLI2022108781}. Compared to traditional coarse-grained sentiment analysis, however, training a model to perform ASC requires a dataset with richer annotations.

Despite the growing number of finance-specific ASC datasets made public (e.g., SemEval 2017 Task 5, FiQA, SEntiFiN 1.0), the majority of literature is built upon English corpora, excluding low-resource languages from the scope of research \cite{cortis-etal-2017-semeval, 10.1145/3184558.3192301, RePEc:bla:jinfst:v:73:y:2022:i:9:p:1314-1335}. Aiming to facilitate finance-specific ASC research using the Korean language we present \( KorFinASC\), a Korean aspect-level sentiment classification dataset for finance consisting of 12,613 human-annotated samples. \( KorFinASC\) is, to the best of our knowledge, the largest publicly available human-generated finance-specific dataset and aspect-level classification dataset, written in Korean, each irrespective of task and domain.
\begin{figure}
  \centering
  \includegraphics[width=\linewidth]{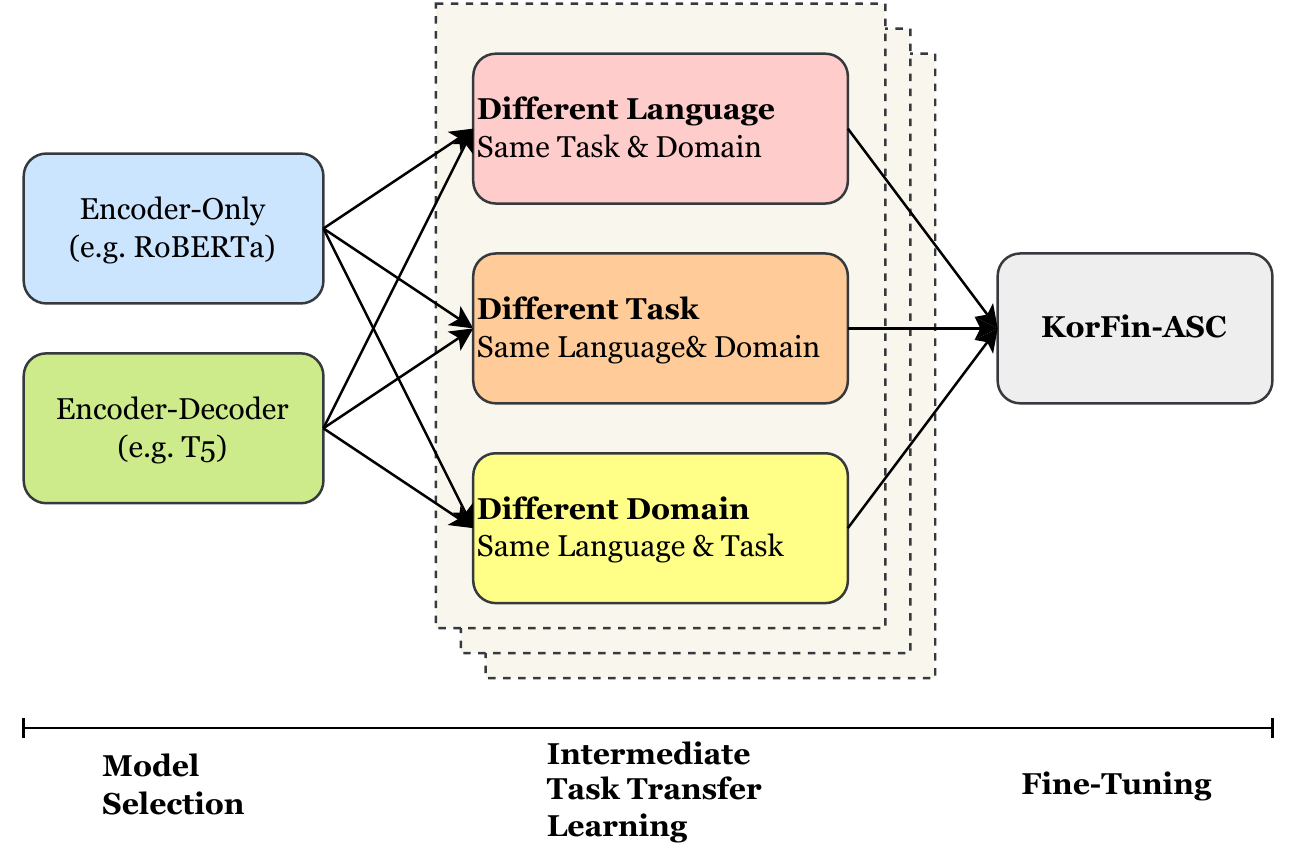}
  \caption{Experimental pipeline with variants of pre-trained language models, and intermediate tasks for intermediate transfer learning. 
  \label{figure:mtlb_1}
}
\end{figure}
Further in this work, we use the pipeline in figure~\ref{figure:mtlb_1} over \(KorFinASC\) to offer a comprehensive investigation into whether datasets of different domains, languages, and tasks can be leveraged to improve classification performance. Also, referred to as intermediate transfer learning, multiple researches has already reported that cross-lingual transfer from English improves performance in general domain tasks \cite{pruksachatkun2020intermediate,sattar2021multi}. We find that the nature of finance-specific ASC problems differs significantly from that of general domain tasks requiring additional measures to be properly transferred. We postulate that such behavior origins from ``non-stationary knowledge'', once right facts that shift with time. Modern-day natural language processing (NLP) has been built upon the assumption that the likelihood of a word's appearance is constant over time. However, financial markets are extremely non-stationary systems, where exploiting past knowledge may lead to a lack of robustness \cite{schmitt2013non}. Our work provides empirical evidence that repeated appearances of financial entities in pre-training, intermediate training, and fine-tuning deeply encode non-stationary knowledge regarding financial entities in pre-trained language models (PLM), hence overestimating its predictive ability on test datasets collected from the same period of time. To mitigate such biases from distorting predictions we introduce ``TGT-Masking'', a novel masking pattern that restricts PLMs from speculating non-stationary knowledge. We discover that combined with a series of transfer learning, ``TGT-Masking'' improves 22.63\% of classification accuracy compared to standalone models on \( KorFinASC\). 

\noindent In summary, our contributions are as follows: 

\begin{itemize}
\item We create and make public \( KorFinASC\), the first finance-specific ASC dataset in Korean.
\item We report the existence of ``non-stationary knowledge'' that has overestimated the classification accuracy of PLMs and introduce ``TGT-Masking" a novel masking pattern to restrict PLMs from speculating ``non-stationary knowledge."
\item We discover that datasets of different domains, languages, and tasks can be leveraged to improve classification performance for low-resource languages on finance-specific ASC tasks.
\end{itemize}

\section{Background and Related Work}

\subsection{Aspect-Level Sentiment Classification in Finance}
Ever since it has been proven that textual data is correlated with stock price, a growing number of researchers have devised methods to enhance the predictive power via sophisticated NLP algorithms \cite{wysocki1998cheap}. Among the various attempts, the largest branch of research employs sentiment analysis to translate financial corpora into sentiment signals that enhance investment decision-making \cite{10.2307/20122297, ding-etal-2014-using}. Building upon past literature, the most recent work attempt to better capture sentiment from sentences with complex syntactic structures by introducing aspect-level sentiment classification (ASC) \cite{CONSOLI2022108781}. In contrast to traditional course-grained sentiment analysis, which is incapable of disambiguation of sentences with multiple sentiments, ASC involves the classification of sentiment with regard to each aspect present in the input. ASC, due to its relatively short history, has not been fully explored in finance, nonetheless, it is anticipated by academia to improve sentiment signal extraction by filtering out noises irrelevant to the target in interest \cite{RePEc:bla:jinfst:v:73:y:2022:i:9:p:1314-1335}.

\subsection{Intermediate Transfer-Learning}
Unlike traditional learning, in which datasets and training are strictly isolated for each task, transfer learning involves using datasets from different domains, languages, and tasks to improve performance in a target task. For instance, given source and target domains \(D_s\) and \(D_t\), source and target tasks \(T_s\) and \(T_t\), and source and target languages \(L_s\) and \(L_t\), learning the target conditional probability for target task from a dataset where \(D_s \neq D_t\),  \(T_s \neq T_t\), or  \(L_s \neq L_t\) can be referred to as transfer learning \cite{DBLP:journals/corr/abs-1911-02685}.
The majority of previous research on cross-lingual transfer for general-domain ASC involves parallel annotated data for both the source and target languages; ironically such data is equally difficult to acquire \cite{balahur-turchi-2012-multilingual, lambert-2015-aspect}. Only very recently unsupervised cross-lingual transfer was proposed, and surprisingly it has reported 20 to 23\% improvement on state-of-the-art approaches \cite{sattar2021multi}. However, whether such behavior is equally applicable to the finance domain has not yet been explored. 

\subsection{Financial Biases in Pre-trained Language Models}
Leveraging the vast amount of text corpora available on the internet, PLMs have achieved unprecedented levels of performance, with some even surpassing human baselines \cite{DBLP:journals/corr/abs-1910-10683, brown2020language}. Recently, however, an increasing number of research have pointed out that stereotypical biases in the text corpora have been propagated to PLMs raising questions over the fairness of these models \cite{nadeem-etal-2021-stereoset}. The issue of prejudiced knowledge is equally prevalent in the finance domain, as FinBERT, the most widely used finance-specific PLM model, has been reported to prefer stocks from the Materials and Industrials sector amongst others \cite{chuang-yang-2022-buy}. 

\section{Non-Stationarity in Financial Corpora}
Financial markets are well known to be non-stationary, implying that historical data may not always lead to accurate forecasts \cite{schmitt2013non}. In this paper, we question whether contemporary PLMs, which assume that the likelihood of a word's appearance is static, is suitable for financial decision-making. We hypothesize that the non-stationarity of financial data is likewise represented in financial corpora, making PLMs trained on such corpora equally non-stationary. To demonstrate whether PLMs contain outdated knowledge regarding financial entities we select two PLMs trained on a corpus collected from a different period of time, BERT and FinBERT. For financial entity C, we construct a template sentence “C is a [MASK] company” and through a masked token prediction task probe the two models.

\begin{figure}[ht]
\centering
\includegraphics[width=\columnwidth]{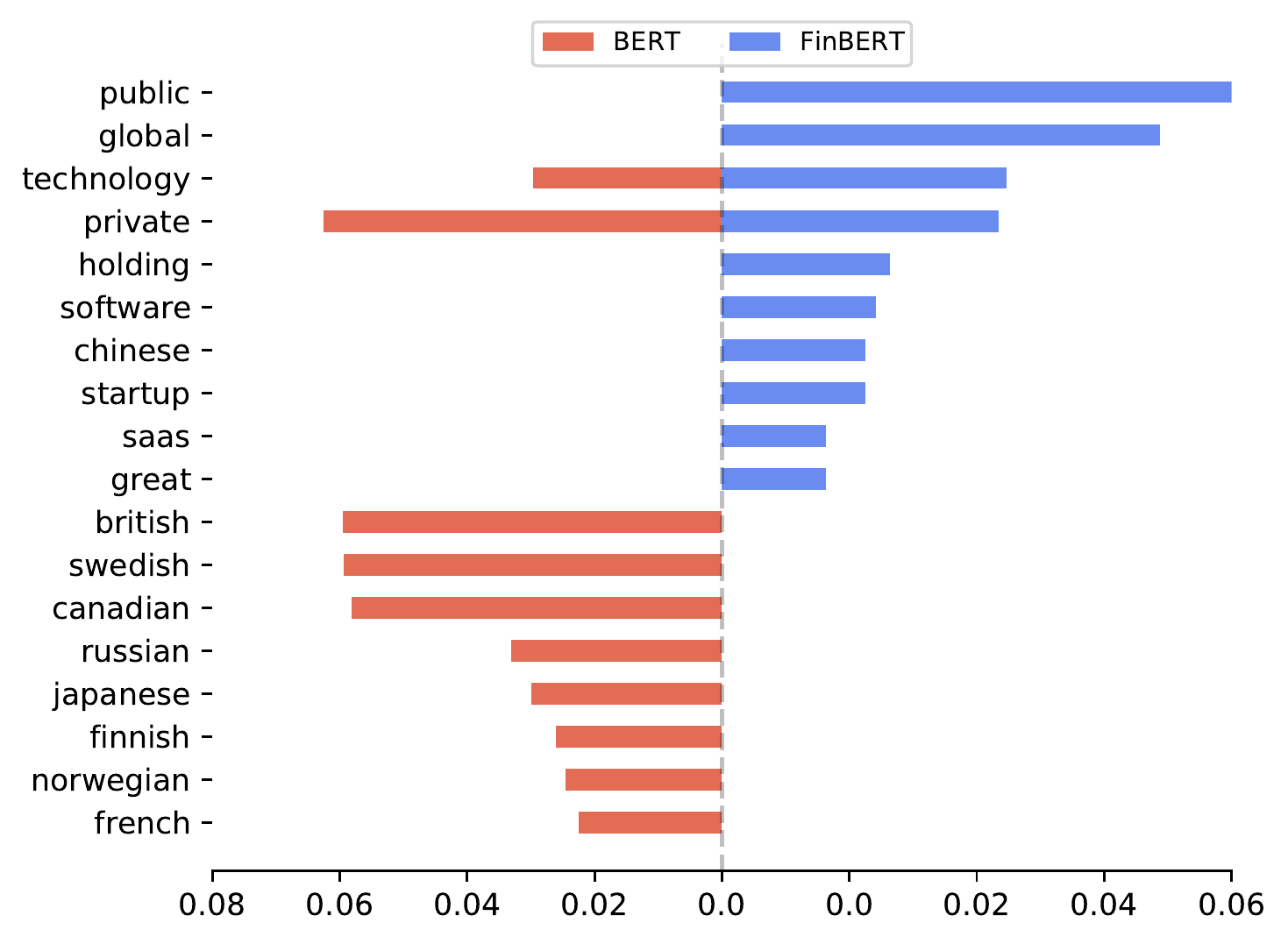}
\label{figure:mtlb_4}
 \caption{The 10 most probable words and their conditional probability predicted by BERT and FinBERT for ``Tesla is a [MASK] company.''}\label{figure:mtlb_2}
\end{figure}

Figure~\ref{figure:mtlb_2} is the 10 most probable words and their conditional probability predicted by BERT and FinBERT using the above-mentioned prompt for companies Tesla, and Ford. Surprisingly, with the exception of a small number of phrases,  the two models predict distinct outputs. We presume that this behavior is due to the difference in the corpora used to train each model. Moreover, we observe that the generated words can be sorted into the following four categories.

\begin{figure}[ht]
\centering
\includegraphics[width=\columnwidth]{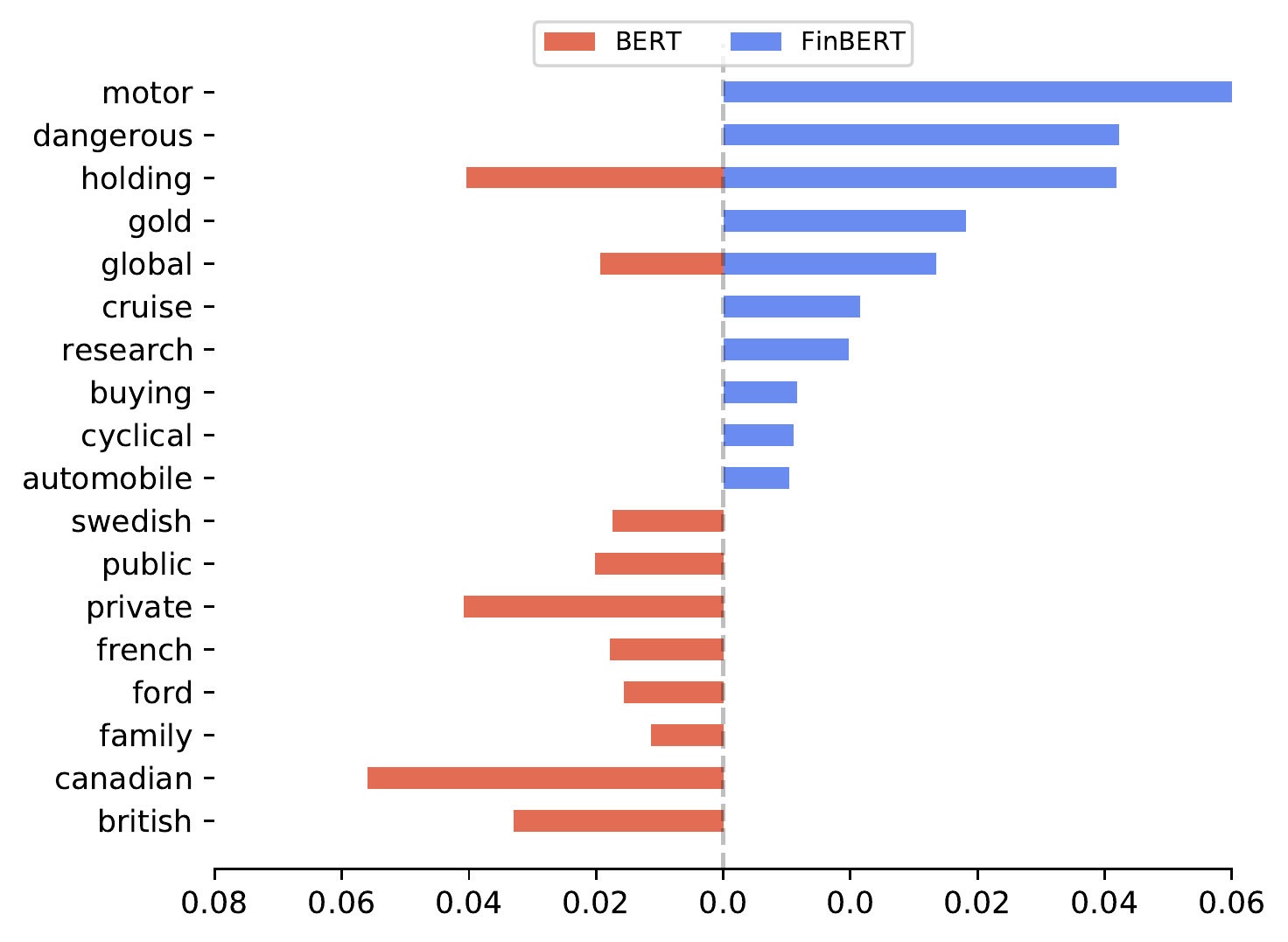}\label{figure:mtlb_5}
 \caption{The 10 most probable words and their conditional probability predicted by BERT and FinBERT for ``Ford is a [MASK] company.''}\label{figure:mtlb_3}
\end{figure}

\begin{enumerate}
\item \textbf{Correct Generations:} 
Words such as ``technology'', and ``motor'' are correct generations that well explain the target entities.
\item \textbf{Wrong Generations: }%
Words such as ``chinese'', ``british'', ``canadian'' or ``gold'' are factually wrong generations that do not fit the target entities. 
\item \textbf{Sentiment-Bearing Generations:}
FinBERT generates ``great'' for Tesla and ``dangerous'' for Ford, both of which are evaluative terms.
\item \textbf{Once Correct Generations:}
Interestingly, we also observe generations that were once correct but outdated. For instance, both models predict the term ``private'' for Tesla, which suggests that Tesla is a private firm. However, Tesla became public in 2010 and has remained so since.
\end{enumerate}

All cases corresponding to categories 2, 3, and 4 are critical if propagated to downstream tasks; however, factually wrong generations have been dealt with often in recent years, so our work focuses on sentiment-bearing and once-correct generations \cite{weidinger2021ethical}.  The two errors are similar in that they incorrectly use historical distributions to forecast future values, i.e., they mistake textual data to be stationary.

The existence of sentiment-bearing generations reaffirms that language models have been trained to prefer a specific entity over the other, even in the absence of context. It is a dangerous assumption for PLMs to associate a financial institution with a certain sentiment based on historical data, given a great quantity of research in finance has already shown that past returns have little or no bearing on future returns \cite{fama1965behavior, huang2020time}. Once-correct generations, to the best of our knowledge, have never been mentioned in previous financial NLP studies. However, the existence of once-correct generations indicates that current PLMs, in which training is ceased once served, will become obsolete as soon as they do so, meaning that they cannot keep up with the non-stationary behavior of financial data. Accordingly, to generically refer to sentiment-bearing and once-correct generations, two errors that mistake text to be non-stationary, we coincide the term “non-stationary knowledge” and demonstrate its presence and toxicity further in our work.

\section{Dataset Creation}
The goal of this research is to explore whether datasets and PLMs of different languages, tasks, and domains, can have their non-stationary knowledge removed and then transferred to improve the performance of finance-specific ASC in low-resource languages. However, finance-specific ASC datasets in languages other than English, are not publicly available. Accordingly, we create \( KorFinASC\), the first finance-specific fine-grained sentiment analysis dataset in Korean.

\begin{table*}[ht]
 \centering{
\begin{tabular}{lllllll}
\hline
             &        &                  & \multicolumn{4}{l}{Entity-Sentiment Annotations}          \\ 
             & Num Sample & Average Word Num & Positive       & Negative       & Neutral        & Total  \\ \hline
SINGLE\_ENT  & 5,964  & 15.6             & 2,491 (41.7\%) & 2,082 (34.9\%) & 1,391 (23.3\%) & 5,964  \\
MULTIPE\_ENT & 2,779  & 18.4             & 2,079 (31.3\%) & 2,532 (30.7\%) & 2,532 (38.1\%) & 6,649  \\
Total        & 8,743  & 16.5             & 4,570 (36.2\%) & 3,923 (32.6\%) & 3,923 (31.1\%) & 12,613 \\ \hline
\end{tabular}%
}
\caption{KorFin-ASC statistics.}
\label{table:mtlb_2}
\end{table*}

\subsection{Data Collection}
Neural models trained to perform sentiment classification on financial statements are expected to behave analogously with a human investor; thus we collect the training data from Naver Finance\footnote{https://finance.naver.com}, an analyst report aggregator commonly used in the finance industry. The collected reports were segmented into sentences and then we applied the following pre-processing methods to remove undesirable context.

\begin{itemize}
\item Remove sentences with less than 40 characters.
\item Remove sentences including phrases such as:  \\ 
All Rights Reserved,
\includegraphics[height=0.25cm]{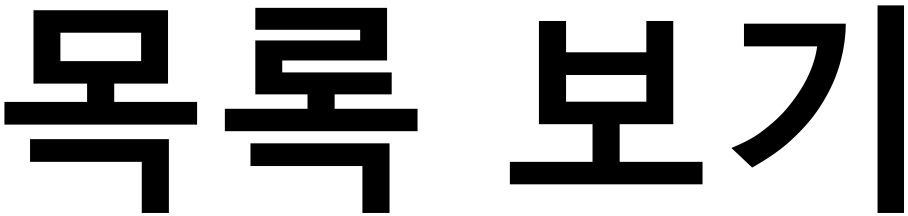} (Click For Contents),  \\
 @domain.com,
\includegraphics[height=0.25cm]{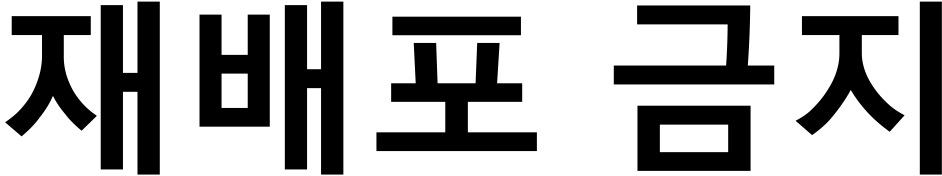} (Redistribution Prohibited)
\item Remove sentences with less than 2 organization entities included.
\end{itemize}

\subsection{Annotation process} 
For the annotation process, three native Korean speakers with degrees in finance and economics were employed. Each annotator was provided with 10,000 partially overlapping samples and was instructed to identify existing entities and annotate the subjected sentiment. The annotators were required to think as real-life investors and assign sentiment between “positive” or “negative” based on how the sample will contribute to the market value of the entity. Once they encounter sentences with insufficient information for judgment they were guided to either label “neutral” or remove the sentence. 
However, difficulties exist to reach a perfect inter-annotator agreement, particularly when more than one annotator is involved in the production of the dataset. To minimize disagreement, the annotators were provided table~\ref{table:mtlb_1} from one of the primary authors for sentiment labeling and were advised to refrain from speculating previous knowledge beyond the given conditions.

To measure the consistency of annotations from different annotators, an agreement study has been conducted using a subset of 611 overlapping samples. Surprisingly only 12 samples, approximately 1.9\%, failed to meet a consensus. The final decision for the above-mentioned samples was provided by the primary authors.

\begin{table}[ht]
\centering{
\begin{tabularx}{\columnwidth}{lX}
 \toprule
Sentiment & Example Cases  \\\midrule
Positive  & Overperformed market expectations/market/sector/competitor, Raised Funding, Analyst buy rating, Entry to new market, Alleviation of risk.            \\
Negative  & Underperformed market expectations/market/sector/competitor, Failed fund raising, Analyst sell rating, Default filing, Employee layoff,  Owner risk.\\ \bottomrule
\end{tabularx}%
}
\caption{Pre-definded cases of Positive/Negative samples provided by the primary authors.}
\label{table:mtlb_1}
\end{table}

\subsection{Dataset Statistics}

Resultingly, we build and release \( KorFinASC\), a Korean aspect-level sentiment classification dataset for finance consisting of 12,613 human annotated samples. To the best of our knowledge, \( KorFinASC\) achieves to be the largest publicly available human-generated finance-specific dataset and aspect-level sentiment classification dataset, written in Korean, each irrespective of task and domain.

\( KorFinASC\) contains 8,743 unique samples annotated with entities and corresponding financial nuances. Each sample is tagged either MULTIPLE\_ENT or SINGLE\_ENT depending on the number of financial entities included. MULTIPLE\_ENT samples make up 31.8\%, contributing 6,649 entity-sentiment annotations to the dataset. The distribution of sentiments was monitored throughout the annotation process resulting in a fairly low-class imbalance as shown in table~\ref{table:mtlb_2}.

However, one feature that concerns the authors is that 4,354 unique entities appear throughout the dataset, indicating an average of 2.9 appearances per each. Accordingly, per-entity sentiment distribution is highly imbalanced for most of the entities which might mislead a neural model trained on this corpus to directly link an entity to a specific sentiment instead of analyzing its semantics.

\section{TGT-Masking}
Earlier in this work, we have criticized that PLMs grounding predictions on non-stationary knowledge, are toxic behaviors likely to overestimate the model's predictive power. To demonstrate that this tendency is a language-agonstic behavior resulting from the nature of PLMs, we conduct a perturbation sensitivity analysis (PSA) on RoBERTa-Large and KLUE-RoBERTa-Large models trained using SentFiN and KorFin-ASC \cite{prabhakaran2019perturbation, liu2019roberta, park2021klue}. PSA measures the sensitivity of a classification model by calculating the mean, and standard deviation of scores with entity-perturbed inputs. For this process, we use a template sentence “[TARGET\_ENTITY\_1] rallied while [TARGET\_ENTITY\_2] falls into a bear market.”, to probe the models. Sentiment scores for each polarity were derived by applying a softmax function on the model’s output layer. To determine the perturbation sensitivity for the positive polarity, perturbations were applied to [TARGET ENTITY 1], and for negative polarity, perturbations were applied to [TARGET ENTITY 2].

\begin{figure}[ht]
\centering
\includegraphics[width=\columnwidth]{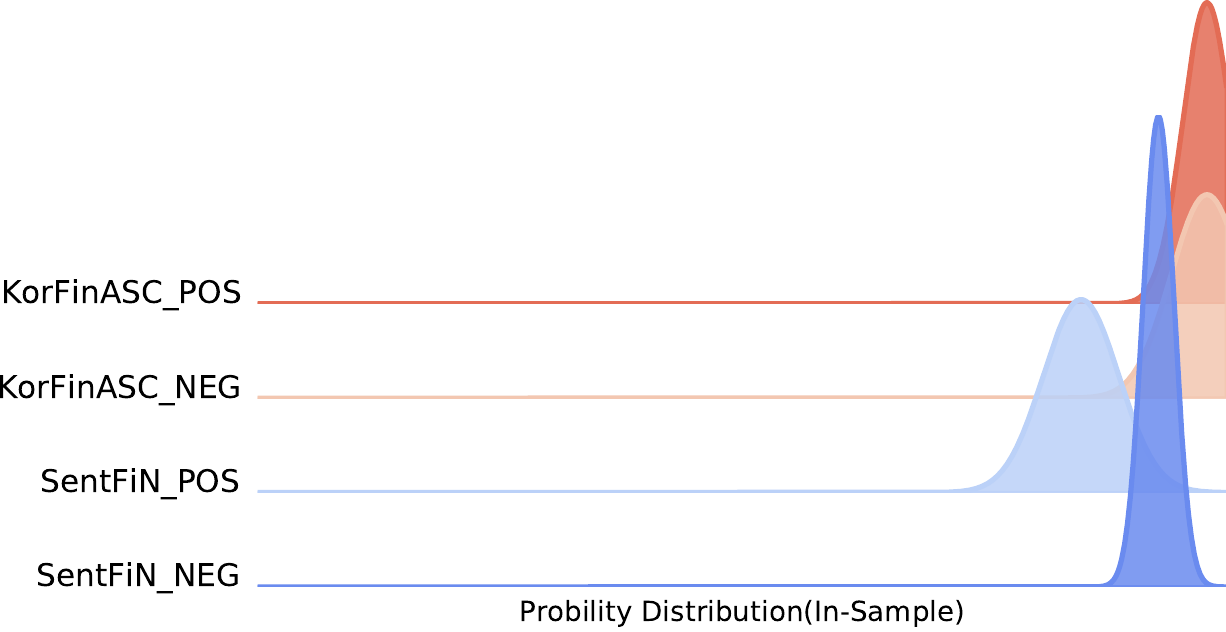}\label{figure:mtlb_6-7}
\includegraphics[width=\columnwidth]{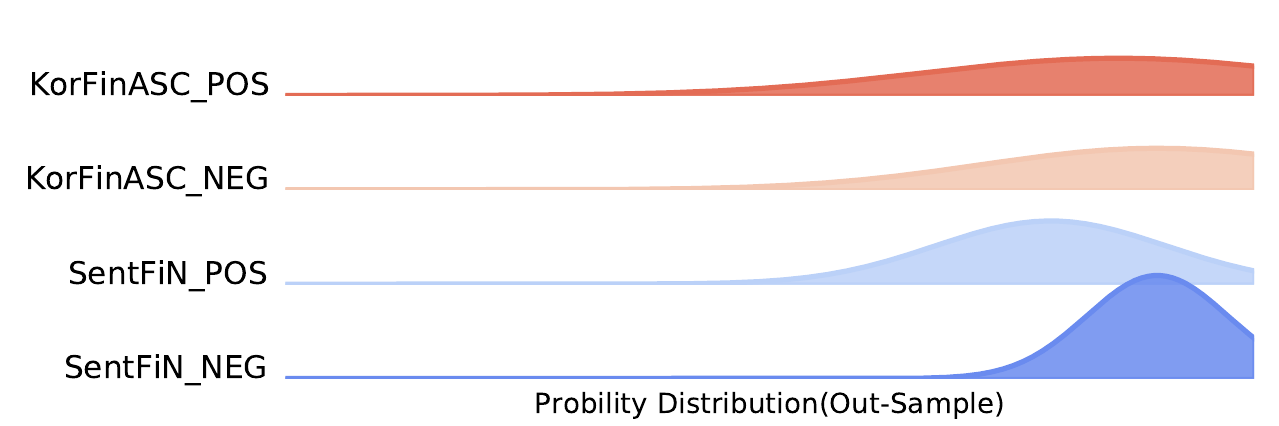}\label{figure:mtlb_8-9}
 \caption{A plot assuming a normal distribution for polarities positive(POS) and negative(NEG), over datasets KorFinASC and SentFiN. }\label{figure:mtlb_6-9}
\end{figure}

Formally, \(X_\eta\) denotes the perturbed sentence with the target entity replaced with an alternative entity \(\eta \in N\). To compare the model's behavior to perturbations from in-sample and out-of-sample entities two types of \(\eta\) are introduced. \(X^i_\eta\) denotes a sentence perturbed with an entity already seen from the training phase while \(X^o_\eta\) denotes a perturbation with an entity never seen by the model before. Finally, \(f^P(X)\) is used to denote the positive polarity score and \(f^N(X)\) for the negative polarity score. The calculation of scores is repeated \(N\) times each for English and Korean.

Ideally, a text classification model must produce scores independent from the financial entities discussed in the context. However, sentences subjected to perturbations resulted in an output with a standard deviation ranging from 0.016 to 0.2.  In figure~\ref{figure:mtlb_6-9}, we observe that models exhibit lower confidence levels and more uncertainty in response to perturbations involving out-of-sample entities (distributions in red). For instance, the average mean score of \(F^P(X)\) and \(F^N(X)\) for out-of-sample entities was 0.06 points lower for both models when compared to sentences with in-sample entity perturbations. Consequently, we discover that PLMs referencing non-stationary information result in unstable prediction ability that varies depending on the financial entity present in the context.

To mitigate the issue we propose “TGT-Masking”, a novel masking pattern that restricts PLMs from speculating non-stationary knowledge. TGT-Masking applies to both the training and test phases by substituting the financial entity of interest with a special token [TGT], a short for ``target''.  Doing so, not only prevents the PLM from learning non-stationary traits about the entity during the training phase but also, by replacing the target entity with an identical token for each prediction, keeps the output polarity score for homogeneous contexts constant. We postulate that the proposed method is a better option for debiasing compared to past approaches which mostly concentrated on de-biasing augmentations or time-aware embedding~\cite{lee2021crossaug, guo-etal-2022-auto, dhingra2022time}.  First, debiasing-augmentation approaches involve using a fixed prompt to generate unseen samples to align the model’s predicted label distribution, inevitably training the model on an identical template with varying entities multiple times. However, recent work has proven that duplication of data harms the model from achieving a higher accuracy \cite{lee2021deduplicating}. Second, augmentation methods require a generative language model to generate data samples requiring large amount of computation resources. Moreover, time-aware embeddings are also a data-heavy solution since they require timestamped data to address the temporal change of knowledge. Compared to previous solutions, TGT-Masking is implemented by simple replace functions in the data loader without requiring additional generative models, computing resources, or datasets which makes it the cheapest solution available.

\section{Experimental Setup}
We conduct our experiments, using the pipeline mentioned in figure~\ref{figure:mtlb_1}, on \( KorFinASC_{test}\) and \( KorFinASC\)-\(Perturb_{test}\). \( KorFinASC\)-\(Perturb_{test}\) is a maliciously modified version of the original test set in which the original financial institutions have been substituted with out-of-distribution entities, similar to control tasks described in past literature\cite{hewitt-liang-2019-designing}. As larger models are expected to be more robust towards biases, or non-stationary knowledge in this research, we compare mT5-Large(1.2B params), XLM-RoBERTa-Large(550M params), and KLUE-RoBERTa-Large (337M params). The size of the three models is based on the original BERT recipe, however, mT5 is an encoder-decoder model having twice as many parameters as its encoder-only counterparts. Note that, the difference in parameter count between encoder-only models comes from the larger vocabulary size used on XLM-RoBERTa-Large.

\subsection{Intermediate Transfer Learning}

In our research, we investigate the following types of intermediate transfer learning. When the source and target domains are denoted as \(D_s\) and \(D_t\), source and target tasks as \(T_s\) and \(T_t\), and source and target language as \(L_s\) and \(L_t\), and dataset as \(DS\):

\begin{itemize}
\item \textbf{Language Transfer (Language.T):} 
Uses \(DS = \{Ds=Dt, Ts=Tt, Ls \neq Lt\}\), to inform the model about the task and domain beforehand. In our paper, we use SentFiN.
\item \textbf{Domain Transfer (Domain.T):}
Uses \(DS = \{Ds \neq Dt, Ts=Tt, Ls=Lt\}\), to inform the model about the task and language beforehand. In our paper, we use Ko-ABSA.
\item \textbf{Task Transfer (Task.T):}
Uses \(DS = \{Ds=Dt, Ts \neq Tt, Ls=Lt\}\), to inform the model about the domain and language beforehand. In our paper, we use Financial PhraseBank, and Ko-FinSA. 
\end{itemize}

\noindent Additionally, we test variations with two successive transfers, including Task.T + Language.T, Domain.T + Language.T, and Task.T + Domain.T. Details for the datasets used for intermediate transfer learning are listed in table~\ref{table:mtlb_3}. In our work, PLMs are trained for only 3 epochs in each phase unlike Domain Adaptive Pretraining (DAPT) or Task Adaptive Pretraining (TAPT) described in past publication~\cite{gururangan2020don}. This reduces the models' repetitive exposure to identical texts.

\begin{table*}[ht]
\centering{%
\begin{tabular}{llllllllll}
\hline \hline
 &
  \multicolumn{1}{l|}{} &
  \multicolumn{4}{l|}{ \( KorFinASC_{test}\)} &
  \multicolumn{4}{l}{\( KorFinASC\)-\(Perturb_{test}\)} \\
TGT-Masking &
  \multicolumn{1}{l|}{Transfer Learning} &
  \multicolumn{2}{l}{SINGLE\_ENT} &
  \multicolumn{2}{l|}{MULTIPLE\_ENT} &
  \multicolumn{2}{l}{SINGLE\_ENT} &
  \multicolumn{2}{l}{MULTIPLE\_ENT} \\
 &
  \multicolumn{1}{l|}{} &
  ACC &
  F1 &
  ACC &
  \multicolumn{1}{l|}{F1} &
  ACC &
  F1 &
  ACC &
  F1 \\ \hline \hline
\multicolumn{10}{l}{Model: XLM-RoBERTa-Large} \\
TRUE &
  \multicolumn{1}{l|}{N/A} &
  77.85 &
  73.24 &
  79.23 &
  \multicolumn{1}{l|}{79.27} &
  77.85 &
  73.24 &
  79.23 &
  79.27 \\
 &
  \multicolumn{1}{l|}{Language.T} &
  79.24 &
  75.74 &
  81.79 &
  \multicolumn{1}{l|}{81.77} &
  79.24 &
  75.74 &
  81.79 &
  81.77 \\
 &
  \multicolumn{1}{l|}{Task.T + Language.T} &
  79.35 &
  76.54 &
  82.91 &
  \multicolumn{1}{l|}{82.58} &
  \textbf{79.35} &
  \textbf{76.54} &
  \textbf{82.91} &
  \textbf{82.58} \\
 &
  \multicolumn{1}{l|}{Language.T + Domain.T} &
  79.12 &
  75.66 &
  82.21 &
  \multicolumn{1}{l|}{82.24} &
  79.12 &
  75.66 &
  82.21 &
  82.24 \\
FALSE &
  \multicolumn{1}{l|}{N/A} &
  79.18 &
  75.35 &
  81.42 &
  \multicolumn{1}{l|}{81.49} &
  65.16 &
  60.62 &
  60.28 &
  58.05 \\
 &
  \multicolumn{1}{l|}{Language.T} &
  79.41 &
  75.3 &
  83.82 &
  \multicolumn{1}{l|}{83.67} &
  61.1 &
  58.24 &
  61.79 &
  59.69 \\
 &
  \multicolumn{1}{l|}{Task.T + Language.T} &
  \textbf{82.87} &
  \textbf{79.13} &
  \textbf{83.67} &
  \multicolumn{1}{l|}{\textbf{83.69}} &
  72.82 &
  68.33 &
  58.47 &
  57.94 \\
 &
  \multicolumn{1}{l|}{Language.T + Domain.T} &
  79 &
  73.99 &
  82.15 &
  \multicolumn{1}{l|}{82.11} &
  62.17 &
  61.74 &
  64.16 &
  61.09 \\ \hline
\multicolumn{10}{l}{Model : mT5-Large} \\
TRUE &
  \multicolumn{1}{l|}{N/A} &
  70.76 &
  64.6 &
  57.86 &
  \multicolumn{1}{l|}{55.41} &
  70.76 &
  64.6 &
  57.86 &
  55.41 \\
 &
  \multicolumn{1}{l|}{Language.T} &
  68.75 &
  \textbf{75.72} &
  60.69 &
  \multicolumn{1}{l|}{58.11} &
  68.75 &
  \textbf{75.72} &
  60.69 &
  58.11 \\
 &
  \multicolumn{1}{l|}{Task.T + Language.T} &
  76.06 &
  70.59 &
  \textbf{79.74} &
  \multicolumn{1}{l|}{\textbf{77.41}} &
  76.06 &
  70.59 &
  \textbf{79.74} &
  77.41 \\
 &
  \multicolumn{1}{l|}{Language.T + Domain.T} &
  72.99 &
  67.77 &
  74.45 &
  \multicolumn{1}{l|}{71.31} &
  72.99 &
  67.77 &
  74.45 &
  71.31 \\
FALSE &
  \multicolumn{1}{l|}{N/A} &
  77.06 &
  71.55 &
  64.11 &
  \multicolumn{1}{l|}{62.31} &
  76.12 &
  69.77 &
  60.64 &
  58.76 \\
 &
  \multicolumn{1}{l|}{Language.T} &
  \textbf{78.24} &
  73.1 &
  79.54 &
  \multicolumn{1}{l|}{77.22} &
  73.38 &
  65 &
  71.07 &
  68.59 \\
 &
  \multicolumn{1}{l|}{Task.T + Language.T} &
  72.77 &
  68.83 &
  79.23 &
  \multicolumn{1}{l|}{76.32} &
  \textbf{76.51} &
  71.76 &
  79.39 &
  \textbf{77.77} \\
 &
  \multicolumn{1}{l|}{Language.T + Domain.T} &
  53.64 &
  48.89 &
  47.43 &
  \multicolumn{1}{l|}{44.03} &
  40.29 &
  36.42 &
  36.04 &
  32.98 \\ \hline
\multicolumn{10}{l}{Model: KLUE-RoBERTa-Large} \\
TRUE &
  \multicolumn{1}{l|}{N/A} &
  81.08 &
  76.7 &
  79.94 &
  \multicolumn{1}{l|}{79.72} &
  \textbf{81.08} &
  76.7 &
  79.94 &
  79.72 \\
 &
  \multicolumn{1}{l|}{Task.T} &
  80.25 &
  77.07 &
  83.07 &
  \multicolumn{1}{l|}{83.73} &
  80.25 &
  \textbf{77.07} &
  \textbf{83.07} &
  \textbf{83.73} \\
 &
  \multicolumn{1}{l|}{Domain.T} &
  80.47 &
  76.77 &
  82.26 &
  \multicolumn{1}{l|}{82.23} &
  80.47 &
  76.77 &
  82.26 &
  82.23 \\
 &
  \multicolumn{1}{l|}{Task.T + Domain.T} &
  79.3 &
  76.72 &
  83.01 &
  \multicolumn{1}{l|}{82.41} &
  79.3 &
  76.72 &
  83.01 &
  82.41 \\
 &
  \multicolumn{1}{l|}{Domain.T + Task.T} &
  79.41 &
  76.94 &
  82.51 &
  \multicolumn{1}{l|}{82.03} &
  79.41 &
  76.94 &
  82.51 &
  82.03 \\
FALSE &
  \multicolumn{1}{l|}{N/A} &
  81.98 &
  79.06 &
  83.72 &
  \multicolumn{1}{l|}{83.52} &
  76.62 &
  70.92 &
  77.17 &
  76.59 \\
 &
  \multicolumn{1}{l|}{Task.T} &
  \textbf{82.09} &
  \textbf{79.30} &
  \textbf{85.43} &
  \multicolumn{1}{l|}{\textbf{85.18}} &
  78.4 &
  72.77 &
  58.71 &
  57.88 \\
 &
  \multicolumn{1}{l|}{Domain.T} &
  80.41 &
  76.31 &
  85.13 &
  \multicolumn{1}{l|}{85.05} &
  63.72 &
  61.26 &
  74.13 &
  72.22 \\
 &
  \multicolumn{1}{l|}{Task.T + Domain.T} &
  81.08 &
  78.93 &
  83.62 &
  \multicolumn{1}{l|}{83.39} &
  71.37 &
  69.6 &
  70.26 &
  68.39 \\
 &
  \multicolumn{1}{l|}{Domain.T + Task.T} &
  80 &
  75.63 &
  80.42 &
  \multicolumn{1}{l|}{80.5} &
  52.68 &
  44.81 &
  41.84 &
  43.36 \\ \hline \hline
\end{tabular}%
}
\caption{Results for XLM-RoBERTa-Large, mT5-Large, and KLUE-RoBERTa-Large over  \( KorFinASC_{test}\) and \( KorFinASC\)-\(Perturb_{test}\). All models are trained for 3 epochs in each training phase. F1(MAX) and accuracy are reported for the tests.}
\label{table:mtlb_4}
\end{table*}

\begin{table}[ht]
\resizebox{\columnwidth}{!}{%
\begin{tabular}{lllll}
\hline
Dataset              & \(L_{s}\) & \(D_{s}\)  & \(T_{s}\) & N \\ \hline
SentFiN              & En  & Finance & ASC  & 14k      \\
Financial PhraseBank & En  & Finance & SA   & 5k       \\
Ko-ABSA              & Ko  & General & ASC  & 5k       \\
Ko-FinSA             & Ko  & Finance & SA   & 10k      \\ \hline
\end{tabular}%
}
\caption{Datasets used for intermediate transfer learning. N denotes the number of samples in the dataset.}
\label{table:mtlb_3}
\end{table}

\subsection{Task Formulation}
Unlike conventional sentiment analysis tasks which receive raw sentences as input, target aspects must be specified for ASC tasks. For models which use TGT-Masking during its training phase, we anticipate the existence of [TGT] mask to guide the model to analyze the input in the interest of the token while preventing the encoding of non-stationary information. On the other hand, for models without TGT-Masking, we follow previous research and present the target aspect with a special token delimiting it from the original sequence. Following are example inputs for each condition.

\begin{itemize}
\item \textbf{With TGT-Masking:} \\ 
Energy stocks rallied while [TGT] falls into a bear market.
\item \textbf{Without TGT-Masking:} \\
Energy stocks rallied while S\&P 500 falls into a bear market. [SEP] S\&P 500
\end{itemize}

\subsection{Optimization}
In our work, AdamW with the following parameters was used to optimize the models: r=3e-4(for mT5), 3e-6(for else), betas=(0.9, 0.999), eps=1e-08, weight decay=0.01. Models are trained up to 3 epochs for each training phase, with a maximum length of 300 tokens.

\section{Results}
This section explores and measures the impact of TGT-Masking and intermediate transfer learning on PLMs. F1(MAX) and accuracy are reported separately for SINGLE\_ENT and MULTIPLE\_ENT. SINGLE\_ENT scores represent classification performance for sentences containing a single sentiment. MULTIPLE\_ENT denotes scores for classification performance over inputs with two or more sentiments included.

\subsection{TGT-Masking and Intermediate Transfer Learning}
Table~\ref{table:mtlb_4} compares XLM-RoBERTa-Large, mT5-Large, and KLUE-RoBERTa-Large over \( KorFinASC_{test}\) and \( KorFinASC\)-\(Perturb_{test}\). Surprisingly, for XLM-RoBERTa-Large, mT5-Large, and KLUE-RoBERTa-Large, models trained with TGT-Masking and intermediate transfer learning overperform their standalone versions on MULTIPLE\_ENT accuracy by 22.63, 19.1, and 5.9 percentage points, correspondingly. Furthermore, through our experiments, we report 3 notable findings. 

First, we show that TGT-Masking is an effective way for stopping PLMs from speculating non-stationary knowledge regarding financial entities. Earlier in our work, we discussed that the existence of non-stationary knowledge disturbs PLMs from making reliable predictions independent of the entities given in the input context. In our experiments, models trained with TGT-Masking demonstrate stable performance across both test sets, but models without TGT-Masking underperform themselves on  \( KorFinASC\)-\(Perturb_{test}\). Interestingly, mT5-Large with Task.T + Language.T applied results in greater performance on \( KorFinASC\)-\(Perturb_{test}\); nevertheless, this may also be seen as a failure of PLMs, since ideally, scores should not change and moreover, the improvement cannot be explained.

Second, we provide evidence that intermediate transfer learning improves the performance of PLMs on KorFin-ASC. The performance of the transfer-learned multilingual models is comparable to that of KLUE-RoBERTa-Large, a PLM native to the Korean language, recommending the use of intermediate transfer learning with multilingual models in future research on languages without a native PLM.

Finally, we find that for multilingual models XLM-RoBERTa-Large and mT5-Large, Task.T is preferable to Domain.T. We assume that this is because acquiring knowledge in the finance domain is substantially more challenging than acquiring task-related knowledge. Consequently, it is unlikely for such behavior to be universal over different tasks and domains.

\subsection{Transferring in Lower-Resource Environments}
In the previous section, models were trained using \( KorFinASC\). However, it is difficult to find datasets with \( n>10K\) samples for finance-specific ASC for the majority of languages. Accordingly, we constructed \(KorFinASC(0.3)\), a variation of the original dataset that is 30\% in size. Using \( KorFinASC(0.3)\) we investigate the effectiveness of intermediate transfer learning and TGT-Masking in even lower-resource environments.
The scope of experiments in this section was downsized according to the findings from the preceding section. First, because we have shown that TGT-Masking effectively eliminates non-stationary information from distorting the predictions, we no longer test models without TGT-Masking. Second, for each model, just one transfer learning setting, the one with the highest performance in previous tests, was investigated.

\begin{table}[ht]
\resizebox{\columnwidth}{!}{%
\begin{tabular}{lllll}
\hline \hline
\multicolumn{1}{l|}{}                    & \multicolumn{4}{l}{\( KorFinASC(0.3)_{test}\)} \\
\multicolumn{1}{l|}{Transfer Learning} & \multicolumn{2}{l}{SINGLE\_ENT} & \multicolumn{2}{l}{MULTIPLE\_ENT} \\
\multicolumn{1}{l|}{}                    & ACC     & F1      & ACC    & F1     \\ \hline \hline
\multicolumn{5}{l}{Model: XLM-RoBERTa-Large}                                   \\
\multicolumn{1}{l|}{N/A}                 & 47.18   & 36.13   & 33.58  & 31.25  \\
\multicolumn{1}{l|}{Task.T + Language.T} & 78.47   & 75.54   & 79.21  & 79.26  \\ \hline
\multicolumn{5}{l}{Model: mT5-Large}                                           \\
\multicolumn{1}{l|}{N/A}                 & 55.52   & 50.33   & 52.67  & 45.92  \\
\multicolumn{1}{l|}{Task.T + Language.T} & 76.9    & 72.69   & 78.02  & 75.63  \\ \hline
\multicolumn{5}{l}{Model: KLUE-RoBERTa-Large}                                  \\
\multicolumn{1}{l|}{N/A}                 & 79.02   & 76.23   & 82.36  & 82.31  \\
\multicolumn{1}{l|}{Task.T}              & 77.06   & 73.09   & 79.74  & 79.91  \\ \hline \hline
\end{tabular}
}
\caption{Results for XLM-RoBERTa-Large, mT5-Large, and KLUE-RoBERTa-Large over \( KorFinASC(0.3)_{test}\). All models are trained for 3 epochs in each training phase. F1(MAX) and accuracy are reported for the tests.}
\label{table:mtlb_5}
\end{table}

We observe that test results, table~\ref{table:mtlb_5}, reaffirm our findings. Despite the limited data, intermediate transfer learning allows multilingual models XLM-RoBERTa-Large and mT5-Large, to reach performance that does not differ significantly from previous experiments using full-scale data.  

\section{Conclusion}

Our work explores non-English aspect-level sentiment classification in the finance domain. We discuss ``non-stationary knowledge'' existent in financial corpora that have overestimated the performance of past models and present ``TGT-Masking'', a novel masking pattern, that allows PLMs to output reliable predictions independent from the aforementioned biases. Finally, we investigate a series of intermediate transfer learning and conclude that PLM and datasets from different languages, domains, or tasks can be used to improve classification performance in low-resource languages. We benchmark our findings on KorFin-ASC, the first finance-specific ASC dataset in Korean, and achieve 22.63, 19.1, and 5.9 percentage points improvements with XLM-RoBERTa-Large, mT5-Large, and KLUE-RoBERTa-Large against their standalone counterparts. We release our models, dataset, and code \footnote{Our models and code are available at: \\ https://github.com/guijinSON/KorFin\_ASC}.
\bibliography{aaai23}

\end{document}